\documentclass[conference]{IEEEtran}
\usepackage{cite}
\usepackage{amsmath,amssymb}
\usepackage{graphicx}
\usepackage{booktabs}
\usepackage{array}
\usepackage{url}
\usepackage{hyperref}
\usepackage{listings}
\usepackage{xcolor}
\usepackage{balance}
\usepackage{xspace}
\usepackage{tabularx}
\usepackage{subfig}
\usepackage{graphicx}
\usepackage{enumitem}
\lstdefinestyle{bnf}{
  basicstyle=\scriptsize\ttfamily,
  columns=flexible,
  keepspaces=true,
  breaklines=true,
  frame=single,
  framesep=2pt,
  xleftmargin=0pt,
  xrightmargin=0pt,
  aboveskip=1pt,
  belowskip=1pt,
  backgroundcolor=\color{gray!8},
}
\newcommand{\pegasusAI}{HAIDL-X\xspace}
\title{Human--AI Interactions as First-Class Data:\\
A Formal Interaction Model and Runtime for Explainable AI Collaboration}

\author{
  \IEEEauthorblockN{Michela Taufer}
  \IEEEauthorblockA{University of Tennessee\\
  Knoxville, USA\\
  mtaufer@utk.edu}
  \and
  \IEEEauthorblockN{Kin Wai NG}
  \IEEEauthorblockA{University of Tennessee\\
  Knoxville, USA\\
  kng2@utk.edu}
  \and
  \IEEEauthorblockN{Ian Lumsden}
  \IEEEauthorblockA{University of Tennessee\\
  Knoxville, USA\\
  ilumsden@vols.utk.edu}
  \and
  \IEEEauthorblockN{Aashish Pandey}
  \IEEEauthorblockA{University of Tennessee\\
  Knoxville, USA\\
  apande20@utk.edu}
  \and
  \IEEEauthorblockN{Rajasi Gore}
  \IEEEauthorblockA{University of Tennessee\\
  Knoxville, USA\\
  rgore4@utk.edu}
  \and
  \IEEEauthorblockN{Angelina Ju}
  \IEEEauthorblockA{University of Tennessee\\
  Knoxville, USA\\
  aju@vols.utk.edu}
  \and
  \IEEEauthorblockN{Sukanya Moudgalya}
  \IEEEauthorblockA{University of Tennessee\\
  Knoxville, USA\\
  smoudgal@utk.edu}
  \and
  \IEEEauthorblockN{Ewa Deelman}
  \IEEEauthorblockA{University of Southern California\\
  Marina del Rey, CA, USA\\
  deelman@isi.edu}
  \and
  \IEEEauthorblockN{Sai Swaminathan}
  \IEEEauthorblockA{University of Tennessee\\
  Knoxville, USA\\
  sai@utk.edu}
}

\begin{document}
\maketitle
\begin{abstract}
AI assistants are increasingly embedded in computational notebooks, scientific workflows, software engineering tools, and learning systems, yet human–AI interactions are often treated as transient prompt–response exchanges rather than durable data products, hindering reproducibility, explainability, replay, behavioral analysis, and adaptive AI support.
We argue that these interactions form a new data class requiring formal representation and introduce the Human–AI Interaction Description Language (HAIDL), a formal interaction model with a Backus–Naur Form-inspired concrete syntax.
HAIDL encodes each interaction as a structured object capturing computational context, user request, prompt payload, AI response, feedback, problem context, provenance, and model metadata. We integrate this vision into \pegasusAI, a middleware for JupyterLab that instantiates HAIDL by capturing user-bot state, building contextualized prompts, interfacing with AI providers, collecting feedback, and storing interactions in a repository.
We add a learning layer over HAIDL that maps interaction objects and notebook execution traces into reproducible measures of user behavior, including engagement, performance, help-seeking, error-oriented use, and learning personas.
\pegasusAI provides a foundation for treating human–AI collaboration as structured, queryable, replayable, analyzable data.
\end{abstract}
\section{Introduction}
\label{sec:introduction}
Foundation models are being woven into computational work at every scale: they draft and debug code in computational notebooks, steer data analyses, configure scientific workflows~\cite{pegasus-wms}, and act as agents in campaigns spanning edge, cloud, and HPC systems~\cite{prov-agent}. Computational results are therefore no longer produced by humans or machines alone but through sustained human--AI exchanges, which now shape what gets computed, how errors are resolved, and which conclusions are drawn.
Consider an AI assistant embedded in a computational notebook. A user who never invokes it looks, in the AI logs, like a failure of adoption---but zero interactions are ambiguous: the user may be progressing independently, or stuck and never seeking help, and the logs cannot tell these apart though they call for opposite responses. Heavy use is just as ambiguous: two users may each invoke the assistant a dozen times with nearly identical transcripts, where one is debugging---each request follows a runtime error, each response followed by a code change and a successful re-execution---while the other extracts answers, pasting responses forward without attempts in between. Counted as prompts, these users are identical; as collaborations, they could not be more different. Human--AI collaboration is not \emph{explainable}: when AI assistance fails, misleads, or simply goes unused, we cannot reconstruct why. This is not for lack of logging---emerging telemetry standards record, in structured form, what the model saw and said: prompts, responses, token counts, tool calls, and latencies~\cite{opentelemetrygenai}---but such records cannot answer what an explanation requires: (1)~what computational state led a user to request assistance; (2)~what role that assistance played (error recovery, a next step, a concept, validation); (3)~across a community of users, who is helped, who over-relies, and who struggles in silence. Nor can such records support auditing or reproducing AI-assisted portions of a result after the fact. Answering these requires recording the \emph{human} side of each exchange---the computational state that motivated it and the actions that followed; today's logs do neither.
Existing approaches each capture one side of this coupling (Section~\ref{sec:related_work} surveys these in detail): observability conventions such as OpenTelemetry GenAI~\cite{opentelemetrygenai} instrument the model side but omit user context and downstream actions; provenance models, from W3C~PROV~\cite{w3c-prov} to PROV-AGENT~\cite{prov-agent} and LLMography~\cite{llmography}, trace agent actions within workflows but do not treat the request--response--assessment cycle as a first-class unit; and programming-process formats such as ProgSnap2~\cite{progsnap2} standardize the human side but predate AI assistance, while transcript-only studies~\cite{llm-intents} lack the computational state needed to ground inferred intent. No existing representation couples both sides of a human--AI exchange or defines semantics for deriving knowledge from it.
We provide that representation: the \textbf{Human--AI Interaction Description Language (HAIDL)}, a formal language that models each human--AI exchange as a single structured object spanning the computational state that motivated the request, the contextualized prompt, the AI response, the user's assessment, and full provenance. HAIDL pairs this grammar with an explicit derivation relation ($\Rightarrow_K$) that turns stored interactions into inspectable knowledge---for example, resolving whether a user with zero interactions was progressing independently or was stuck and never sought help. We instantiate HAIDL in \pegasusAI, a human--AI middleware for JupyterLab, and validate it through a semester-long capture of interactions in a university computing course: 35 students, four programming assignments, $\approx$800 recorded interactions, and 140 student--assignment pairs, reproducing the ambiguities above in real data.
This paper makes four contributions:
\begin{itemize}[leftmargin=*,itemsep=1pt,topsep=1pt]
\item \textbf{The HAIDL representation.} A formal model of human--AI interactions as first-class, self-contained data objects, defined by a BNF-inspired grammar over seven semantic components---computational context, request, prompt, response, feedback, problem context, and model provenance---with a derivation relation separating observed data from inferred knowledge (Section~\ref{sec:haidl}).
\item \textbf{A human--AI middleware for runtime capture.} \pegasusAI instantiates HAIDL during live computational work: capturing notebook state per request, constructing contextualized prompts, communicating with interchangeable AI providers, collecting feedback, and persisting interaction objects in a queryable repository (Section~\ref{sec:runtime}).
\item \textbf{Knowledge extraction, demonstrated in a case study.} We define derivation rules composing HAIDL objects with observed execution evidence to recover the role each request played, the trajectory of use over an assignment, and the collaboration patterns behind it---resolving the opening ambiguities: among 37 student--assignment pairs with no recorded interactions, distinguishing students progressing alone from those stuck and never asking (Sections~\ref{sec:knowledge}--\ref{sec:case_study}).
\item \textbf{Empirical quantification of system cost.} We measure the latency, storage, and retrieval overhead \pegasusAI adds to an AI exchange, showing the cost of turning a transient conversation into a persistent, explainable record (Section~\ref{sec:capture_cost}).
\end{itemize}
\section{Human--AI Interactions as First-Class Data}
\label{sec:interaction_data}
\begin{figure*}[t]
  \centering
  \includegraphics[width=0.68\linewidth]{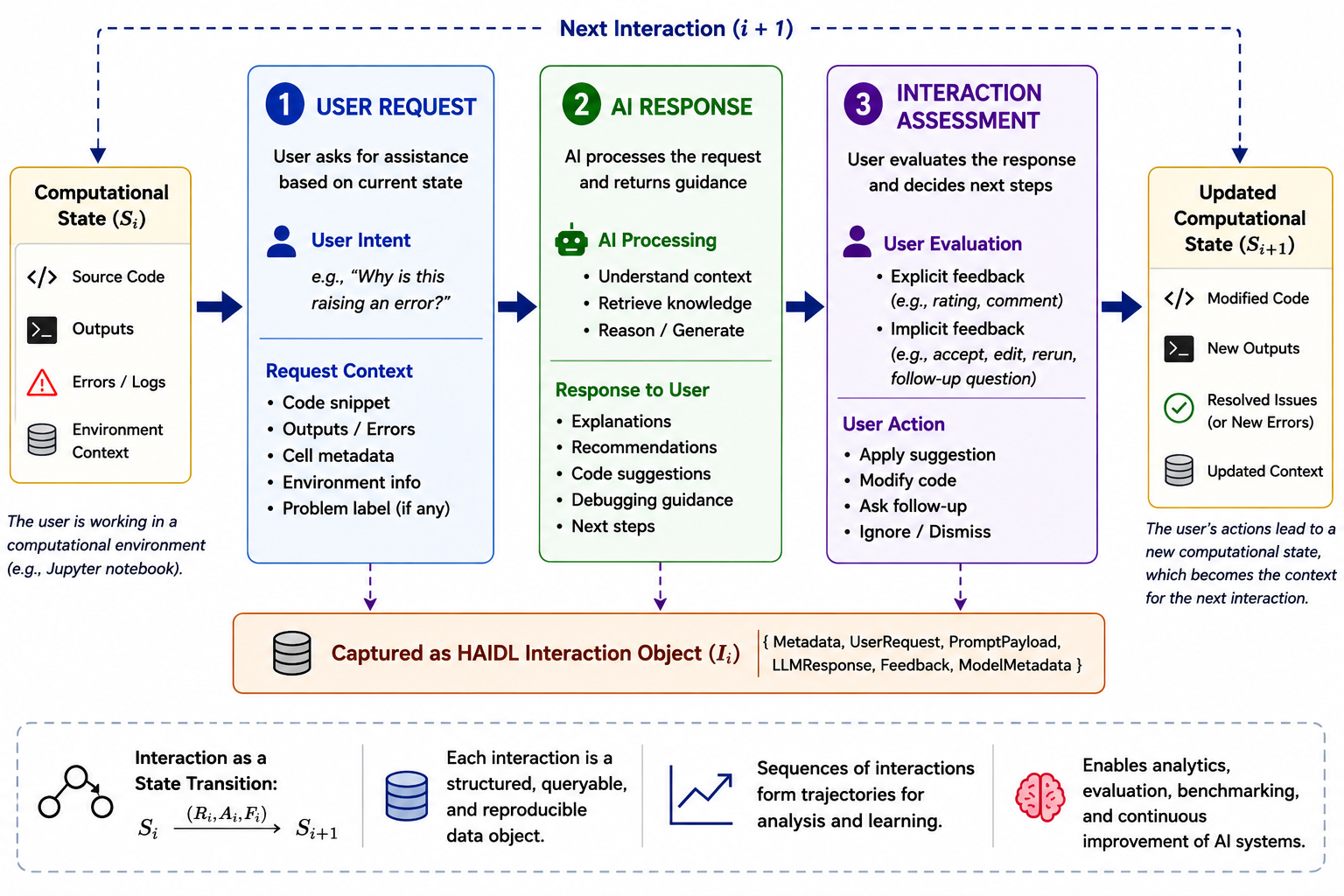}
\caption{Lifecycle of a human--AI interaction. A user request arises from a computational state, the AI returns guidance, and the user evaluates or acts on it. HAIDL represents the full lifecycle as a structured artifact whose outcome seeds the next interaction.}
  \label{fig:interaction_lifecycle}
\end{figure*}

Human--AI collaboration is often seen as a series of prompt--response exchanges, but this view is incomplete. Each interaction is a state-changing event in which a user, an AI system, and a computational environment jointly solve problems: it starts from a computational state that motivates the request and ends in a new state shaped by the user's actions after interpreting the AI's guidance.
We model an interaction as a state transition:
\begin{equation}
I_i : S_i \xrightarrow{(R_i,A_i,F_i)} S_{i+1},
\end{equation}
where \(S_i\) is the computational state before the interaction, \(R_i\) the user request, \(A_i\) the AI response, \(F_i\) the user's assessment of that response, and \(S_{i+1}\) the updated state. Unlike traditional chat logs, this explicitly models how interactions transform computation rather than recording exchanged messages. A sequence of interactions forms an interaction trajectory,
\begin{equation}
T=\langle I_1,I_2,\ldots,I_n\rangle,
\end{equation}
describing the evolution of a collaborative problem-solving session and supporting analysis of interaction strategies, learning behaviors, and AI effectiveness over time. This is the data problem raised in Section~\ref{sec:introduction}: interactions become structured entities to be represented, stored, queried, replayed, and analyzed, like database records or provenance objects in scientific workflows.
Figure~\ref{fig:interaction_lifecycle} illustrates this lifecycle. Every interaction consists of three conceptual stages.
\begin{itemize}[leftmargin=*,itemsep=0pt,topsep=1pt]
\item \textbf{User Request.}
A user in a computational environment submits a request based on the current state (source code, outputs, errors, and problem description), informing both the immediate question and the broader task.
\item \textbf{AI Response.}
Using this context, the AI constructs a contextualized prompt and returns guidance---explanations, debugging help, code suggestions, or conceptual clarification---emphasizing collaborative reasoning rather than mere text.
\item \textbf{Interaction Assessment.}
The user evaluates the response explicitly (feedback) or implicitly (editing code, running cells, follow-ups, or abandoning the solution). This assessment defines the next computational state and the starting point for the next interaction.
\end{itemize}
\section{Human--AI Interaction Description Language}
\label{sec:haidl}
We present the \textbf{Human--AI Interaction Description Language} (HAIDL), a formal language for representing human--AI collaboration as structured data and composing those representations into higher-level interaction knowledge. HAIDL is independent of any AI provider, application, or runtime; its concrete syntax is a grammar inspired by Backus--Naur Form (BNF), while its semantics define how interaction constructs relate to computational state and to one another.
The lower part of Figure~\ref{fig:interaction_lifecycle} shows the basic abstraction. Rather than representing the stages of human--AI collaboration as disconnected events, HAIDL groups them into a single interaction object---a self-contained representation of a collaborative episode that can be stored, shared, queried, replayed, and analyzed independently of its original runtime.
\subsection{Design Principles}
HAIDL builds on five principles:
\begin{enumerate}[leftmargin=*,itemsep=0pt,topsep=1pt]
\item \textbf{Completeness}: capture the full lifecycle of an interaction, not only the prompt and response.
\item \textbf{Modularity}: represent interactions through reusable semantic components.
\item \textbf{Explainability}: associate each construct and derived expression with an explicit interpretation.
\item \textbf{Extensibility}: allow new AI models, applications, and metadata without changing the overall language structure.
\item \textbf{Composability}: allow interaction objects and computational states to be related across time so higher-level knowledge can be expressed and derived.
\end{enumerate}
\subsection{Interaction Objects}
At the highest level, HAIDL represents each interaction as a structured
object:
\begin{equation}
I = \langle M, R, P, A, F, C, L \rangle ,
\end{equation}
where \(M\) is record metadata, \(R\) the user request, \(P\) the prompt
payload, \(A\) the AI response, \(F\) feedback or assessment, \(C\) the
problem context, and \(L\) model metadata.
\begin{figure}[t]
\centering
\includegraphics[width=0.65\linewidth]{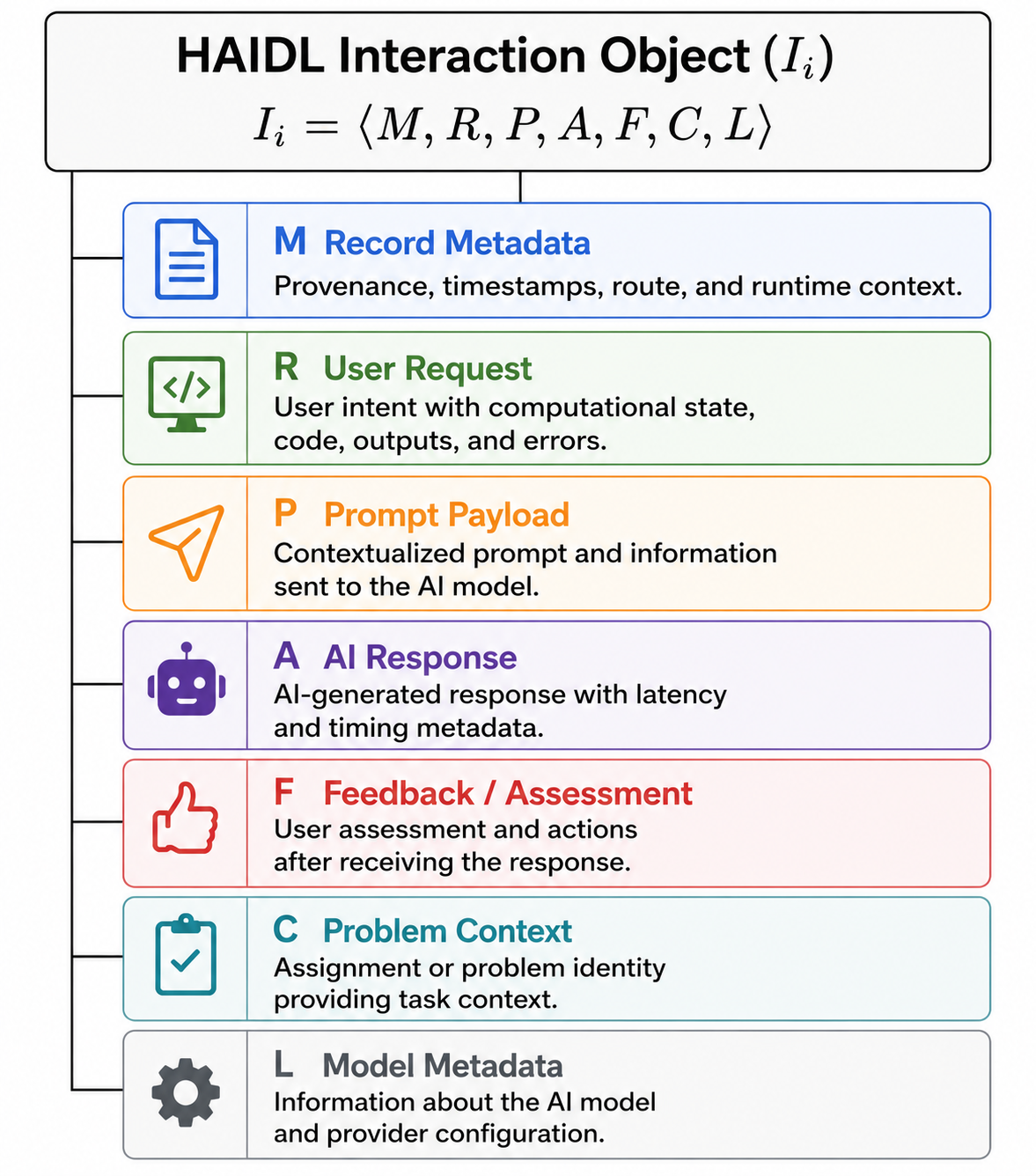}
\caption{Conceptual organization of a HAIDL interaction object: a single
structured object with seven semantic components, forming the atomic unit
of human--AI collaboration and the root production of the HAIDL grammar.}
\label{fig:haidl_object}
\end{figure}
Figure~\ref{fig:haidl_object} shows this organization: rather than storing prompts and responses separately, HAIDL models each collaborative episode as a single object spanning the computational state that led to the request, the contextualized prompt, the AI response, the user's assessment, and the provenance needed to interpret the interaction---the atomic unit from which more complex HAIDL expressions are built. The interaction is:
\begin{lstlisting}[style=bnf,basicstyle=\scriptsize\ttfamily]
Interaction ::= < RecordMetadata, UserRequest, PromptPayload, LLMResponse, Feedback, ProblemContext, ModelMetadata >
\end{lstlisting}
The \texttt{ProblemContext} component links an interaction to the motivating task or scientific problem, which is essential because the same observable interaction can mean different things depending on the problem being solved.
\subsection{Recursive Grammar}
HAIDL recursively decomposes an interaction into semantically meaningful components. For example, \texttt{UserRequest} records the computational context that motivates assistance:
\begin{lstlisting}[style=bnf,basicstyle=\scriptsize\ttfamily]
UserRequest ::= < request_id, user_id, cell_id, Time, code, UserCodeOutputs >
UserCodeOutputs ::= StreamOutput | ErrorOutput
StreamOutput ::= < output_type, traceback, text >
ErrorOutput ::= < output_type, traceback, error_type, error_value >
\end{lstlisting}
Prompt construction, AI response, feedback, problem context, and model metadata are represented separately:
\begin{lstlisting}[style=bnf,basicstyle=\scriptsize\ttfamily]
PromptPayload ::= < prompt, Time >
LLMResponse ::= < request_id, user_id, response, Time, llm_latency, server_latency >
Feedback ::= < request_id, user_id, cell_id, FeedbackValue, Time, problem_label, latency_ms, ts_ms >
FeedbackValue ::= "Yes" | "No" | "Somewhat"
ProblemContext ::= < assignment_name, problem_label >
ModelMetadata ::= < provider, model >
\end{lstlisting}
This decomposition separates information typical chat histories conflate or omit: the computational state motivating assistance, the contextualized prompt sent to the model, the AI response, the user's assessment, the problem context, and runtime and model provenance. Because each component has explicit semantics, they can be queried independently and then recomposed.
\subsection{From Interaction Objects to Interaction Expressions}
The grammar above defines the structure of a single interaction, but human--AI collaboration rarely involves just one exchange. As introduced in Section~\ref{sec:interaction_data} (Eqs.~1--2), an interaction changes computational state and a sequence of interactions forms a trajectory $T=\langle I_1,\ldots,I_n\rangle$. HAIDL supports two composition levels: \emph{structural composition} combines fields into interaction objects per the grammar, while \emph{temporal composition} relates objects via their states, timestamps, users, and problem contexts, expressed at the language level as:
\begin{lstlisting}[style=bnf,basicstyle=\scriptsize\ttfamily]
InteractionTrajectory ::= Interaction*
InteractionData ::=
      Interaction
    | InteractionTrajectory
    | ExecutionState
\end{lstlisting}
An \texttt{Interaction} is a single collaborative episode, an \texttt{InteractionTrajectory} is an ordered sequence of such episodes, and \texttt{ExecutionState} provides observable computational links between them. Together, these form the substrate for expressing higher-level interaction knowledge.
\subsection{From Syntax to Knowledge}
The distinction between representation and knowledge derivation is central to HAIDL. The BNF-inspired production operator \texttt{::=} defines valid language constructs and their structure but does not infer the meaning of observed behavior. We introduce a separate knowledge-derivation relation,
\begin{equation}
O_1,\ldots,O_n \Rightarrow_K K
\label{eq:knowledge_derivation}
\end{equation}
where \(O_1,\ldots,O_n\) are HAIDL interaction-data expressions and \(K\) is knowledge derived from their relationships; \(\Rightarrow_K\) explicitly links observable data to a higher-level interpretation. For example, \texttt{ErrorOutput} appears directly in a \texttt{UserRequest}, so it is interaction data, not inferred knowledge. When that error occurs in the computational state prompting an AI interaction, however, it can support a higher-level interaction expression:
\begin{equation}
\texttt{ErrorOutput}(S_i), I_i
\Rightarrow_K
\texttt{ErrorRecovery}(I_i)
\end{equation}
Similarly, knowledge can be derived from relationships across multiple interactions and computational states:
\begin{equation}
I_i,S_{i+1},I_{i+1}
\Rightarrow_K
K_{i:i+1}
\end{equation}
The exact meaning of \(K_{i:i+1}\) depends on the derivation rule and domain---repeated help seeking, validation of AI guidance, progress following assistance, or another domain-specific pattern. This separation gives HAIDL two complementary roles: (1)~a language for representing and composing interaction data, and (2)~rules for deriving higher-level knowledge from those representations, turning relationships among interactions into inspectable knowledge. In HAIDL, interactions are therefore not only queryable, replayable, and analyzable data objects, but the primitives of a compositional language for human--AI collaboration: its grammar defines syntax, its derivation relation defines semantics, and \pegasusAI provides the runtime that instantiates it from real interactions (Section~\ref{sec:knowledge}).
\section{HAIDL-X: An AI MW for Scientific Computing}
\label{sec:runtime}
HAIDL specifies how human--AI interactions are represented; \pegasusAI is the runtime that realizes this representation as first-class data during execution: a human--AI middleware that augments computational notebooks with context-aware assistance while capturing every interaction as a structured, persistent record. We target JupyterLab~\cite{jupyter}, the interactive computing environment now standard in data science. Rather than a monolithic assistant, \pegasusAI is a layered middleware separating user interaction, interaction capture, AI communication, interaction representation, and persistent storage, letting each layer evolve independently while preserving one common interaction model.
\subsection{Software Architecture}
\pegasusAI comprises six cooperating layers.
\begin{itemize}[leftmargin=*,itemsep=0pt,topsep=1pt]
\item
\textbf{JupyterLab Extension.}
Integrates \pegasusAI into JupyterLab, letting users request assistance and receive responses without leaving the notebook.
\item
\textbf{Frontend Client.}
Captures the computational context for each request---source code, executed cells, outputs, errors, timestamps, notebook metadata---packaging it into structured interaction requests.
\item
\textbf{Backend Server.}
Validates requests, enriches them with context, builds contextualized prompts, invokes external AI providers, and records runtime metadata.
\item
\textbf{HAIDL Interaction Layer.}
Gives the middleware a common representation independent of AI provider or notebook environment; components exchange structured HAIDL objects rather than provider-specific messages.
\item
\textbf{Provider Abstraction Layer.}
Decouples provider-specific APIs from the interaction model, so new foundation models can be integrated without changing HAIDL or the middleware itself.
\item
\textbf{Interaction Repository.}
Persists completed interactions as HAIDL objects in a MongoDB repository, including provenance, timestamps, user assessments, and model information, supporting replay, querying, and behavioral analytics.
\end{itemize}
\subsection{Capture Pipeline, Repository, and Provider Independence}
These six layers execute in sequence for each request: the frontend captures computational state and packages it into a structured request; the backend builds a contextualized prompt and invokes the selected provider through the abstraction layer; the response and context are combined into a HAIDL object with explicit feedback when available; and the completed object is persisted to the repository. This pipeline is what makes human--AI interaction first-class data in practice: instead of ephemeral, implementation-specific chat logs, every interaction becomes a computational artifact---indexed, queried, replayed, and analyzed across sessions, users, assignments, and providers. Because interaction representation is decoupled from model execution, the same HAIDL object results across different AI models.
\section{Deployment and Data Collection}
\label{sec:deployment}
We deployed \pegasusAI in a university Data Engineering course to evaluate HAIDL with interactions from authentic computational work, yielding two complementary views of student activity: HAIDL-based records captured when students interacted with \pegasusAI, and execution evidence preserved in submitted Jupyter notebooks. This section describes the deployment and data preparation; interaction-knowledge derivation is presented in Section~\ref{sec:knowledge}.
\subsection{Classroom Deployment}
\label{sec:classroom_deployment}
\pegasusAI was integrated directly into the JupyterLab environment used by students for programming assignments. When assistance was requested, \pegasusAI captured the active cell code and its available execution output, constructed a context-aware request, routed it to the AI provider, returned the response in the notebook, and persisted the interaction through the HAIDL capture pipeline.
The deployment involved 35 students and four programming assignments, yielding 140 student--assignment pairs. The assignments spanned different tasks and complexity: Assignment~03 focused on mathematical and algorithmic problem solving, Assignment~04 on text processing and exploratory analysis, and Assignments~07 and~08 introduced distributed data-processing tasks using PySpark (distributed $k$-means and a machine-learning pipeline, respectively). \pegasusAI was used during regular assignments, capturing interactions as students wrote, executed, debugged, and inspected notebook code.
\subsection{Experimental Design}
\label{sec:experimental_design}
The classroom deployment used a crossover design. Group~A had access to \pegasusAI for Assignments~03 and~07, while Group~B had access for Assignments~04 and~08. Each group served as the no-access condition for the other assignments, enabling comparisons based on \pegasusAI availability. Assigned access and observed use are not equivalent, however--some students with access never invoked \pegasusAI, while others used it outside the assigned-access condition--so we retain these cases rather than discard them, since the distinction between access and actual use is itself relevant context. The analysis therefore treats the student--assignment pair, not the student alone, as its primary unit.
\subsection{Data Sources}
\label{sec:data_sources}
The deployment produced two complementary data sources:
\paragraph{HAIDL interaction data}
\pegasusAI persisted each observed AI interaction as structured interaction data. Depending on the available context, records include the student identifier, assignment and notebook-cell identifiers, source code, execution outputs or errors, contextualized prompt, AI response, timestamps, model information, latency metadata, and explicit user feedback. These records provide evidence for AI-use measures and semantic interaction events in Section~\ref{sec:knowledge}.
\paragraph{Notebook execution data}
Submitted Jupyter notebooks provide an independent record of computational activity. We analyze notebook cells implementing assignment logic, excluding markdown and timing-only cells. A cell is \emph{touched} when the notebook shows evidence it was executed; among touched cells, one that completes without a Jupyter execution error is successfully executed, one producing an error is unsuccessful, and cells without execution evidence are not executed. These execution-based proxies do not imply a logically correct solution, but provide observable evidence of how far students progressed and how extensively they interacted with an assignment's code cells.
The two sources capture different aspects of the same activity: notebook traces describe what happened during computation, including activity where \pegasusAI was never invoked, while HAIDL objects describe what happened when AI assistance was requested. Neither source alone provides the combined view used in our knowledge derivations.
\subsection{Data Preparation and Alignment}
\label{sec:data_preparation}
We aligned the two data sources around the student--assignment pair: notebook execution records were reduced to the assignment-relevant code cells described above, and \pegasusAI interaction records were grouped by student and assignment and, when a cell identifier was available, linked to the corresponding notebook cell. Within each pair, interactions were ordered by timestamp, preserving the temporal structure HAIDL requires and allowing multiple requests to be represented as a trajectory. The resulting analysis record connects three forms of evidence:
\begin{equation}
D_{u,a} =
\langle
N_{u,a},
H_{u,a},
X_{u,a}
\rangle
\end{equation}
where $N_{u,a}$ denotes notebook execution evidence for student $u$ on assignment $a$, $H_{u,a}$ the corresponding HAIDL interaction objects, and $X_{u,a}$ the assigned \pegasusAI access condition. We preserve these sources separately rather than collapsing them during preprocessing, so that execution behavior, AI interaction behavior, and experimental context remain independently inspectable and are composed only through the explicit derivation rules of Section~\ref{sec:knowledge}.
\section{From Interaction Data to Knowledge}
\label{sec:knowledge}
Representing human–AI interactions as first-class data is useful only if the representation supports reasoning beyond individual records. HAIDL is both a grammar for describing interaction objects and a language for deriving higher-level knowledge; its core idea is composition: observable fields describe individual events, relationships among events describe behavior, and sequences of behaviors reveal higher-level patterns.
We illustrate this in a learning scenario---not to recast HAIDL as a language dedicated to learning tasks, but to show how the interaction primitives of Section~\ref{sec:haidl} combine with domain-specific observations to infer knowledge not explicitly contained in any single interaction record.
\subsection{From Observations to Derived Knowledge}
HAIDL distinguishes three levels of representation:
\begin{lstlisting}[style=bnf]
Observation ::=
      Interaction
    | ExecutionState
InteractionKnowledge ::=
      InteractionEvent
    | InteractionTrajectory
    | BehavioralState
\end{lstlisting}
An \texttt{Observation} is directly captured data. An \texttt{Interaction} is a HAIDL object $I=\langle M,R,P,A,F,C,L\rangle$, and an \texttt{ExecutionState} is the observable notebook state for that interaction (e.g., executed code, outputs, or errors).
\texttt{InteractionKnowledge}, in contrast, is derived by relating observations. For instance, an \texttt{ErrorOutput} is an observation, while using AI assistance in response to that error is an \texttt{ErrorRecovery} event. Likewise, multiple interactions showing persistent lack of progress can form \texttt{RepeatedHelp}. These higher-level constructs are reproducible expressions over HAIDL fields and computational observations, not extra labels stored apart from the interaction data.
We use the derivation operator $\Rightarrow_K$ to distinguish knowledge derived from HAIDL observations from the BNF productions that define their concrete representation:
\[
    O_1,\ldots,O_n \Rightarrow_K K
\]
where $O_i$ are observable interaction or execution objects and $K$ is a derived knowledge construct.
\subsection{Deriving Interaction Events}
\label{sec:interaction_events}
The first level of knowledge describes the apparent role of an AI interaction in the user's computational activity:
\begin{lstlisting}[style=bnf]
InteractionEvent ::=
      ErrorRecovery
    | StuckNextStep
    | ConceptualHelp
    | Validation
\end{lstlisting}
These events are not independent fields in the interaction repository; they are derived from combinations of HAIDL components and notebook state, with intent inferred from the source cell, execution outputs, errors, and the request–response pair.
For example, an error in the computational state for an interaction indicates a need for error-oriented help:
\[
\texttt{ErrorOutput}(S_i), I_i
    \Rightarrow_K
\texttt{ErrorRecovery}(I_i)
\]
An interaction associated with incomplete code or an unfinished computational step provides evidence for:
\[
\texttt{IncompleteState}(S_i), I_i
    \Rightarrow_K
\texttt{StuckNextStep}(I_i)
\]
Requests for explanations or clarification are labeled \texttt{ConceptualHelp}, while requests to inspect or check ongoing work are labeled \texttt{Validation}.
These categories describe \pegasusAI’s observable role in the workflow rather than the user’s underlying intent. Their value lies in being derivable, inspectable, and reproducible from interaction data.
\subsection{Composing Events into Interaction Trajectories}
Individual events show how AI is used at one moment. HAIDL's state-transition model also allows composing events over time. For a learner $u$ and assignment $a$, we define:
\[
T_{u,a}=\langle I_1,I_2,\ldots,I_n\rangle
\]
where interactions are ordered by their timestamps.
\begin{lstlisting}[style=bnf]
InteractionTrajectory ::= InteractionEvent*
\end{lstlisting}
The trajectory permits knowledge to be expressed over relationships among interactions and subsequent computational states. Two useful derived constructs are:
\begin{lstlisting}[style=bnf]
TrajectoryEvent ::=
      RepeatedHelp
    | NoProgress
\end{lstlisting}
Conceptually, repeated help seeking consists of an interaction event followed by a no-progress state and another interaction:
\[
{\scriptsize
\begin{gathered}
\texttt{InteractionEvent} \rightarrow \texttt{NoProgress} \rightarrow \texttt{InteractionEvent} \\
\Rightarrow_K \texttt{RepeatedHelp}
\end{gathered}
}
\]
Likewise, an interaction followed by execution with no observable progress is \texttt{NoProgress}. These constructs highlight a key property of HAIDL: knowledge can be expressed not only over fields within an interaction object, but over relationships among objects and the computational states connecting them.
\subsection{Composing Interaction and Domain Knowledge}
The previous constructs are domain-independent descriptions of interaction
behavior; applications can combine them with domain-specific observations
to derive richer knowledge. In our classroom deployment, notebook
execution provides evidence about student performance and engagement, so
for each student $u$ and assignment $a$, we define:
\[
{\scriptsize
\begin{gathered}
Performance(u,a)=\tfrac{correctly\_solved\_cells(u,a)}{total\_code\_cells(u,a)}, \\
Engagement(u,a)=\tfrac{touched\_cells(u,a)}{total\_code\_cells(u,a)}
\end{gathered}
}
\]
Here, successful execution is used as a performance proxy rather than
proof of semantic correctness. These measures produce four relative
learning profiles:
\begin{lstlisting}[style=bnf]
LearningProfile ::=
      HPHE
    | HPLE
    | LPHE
    | LPLE
HPHE ::= HighPerformance + HighEngagement
HPLE ::= HighPerformance + LowEngagement
LPHE ::= LowPerformance  + HighEngagement
LPLE ::= LowPerformance  + LowEngagement
\end{lstlisting}
Interaction objects provide complementary measures, with $N(u,a)=|\{I_i:user(I_i)=u \land assignment(I_i)=a\}|$:
\[
{\scriptsize
\begin{aligned}
BotIntensity(u,a) &= N(u,a); \\
ErrorUse(u,a) &= \tfrac{|\{I_i:\texttt{ErrorOutput}\in UserRequest(I_i)\}|}{N(u,a)}; \\
FeedbackRate(u,a) &= \tfrac{|\{I_i:Feedback(I_i)\neq\emptyset\}|}{N(u,a)}.
\end{aligned}
}
\]
These measures remain traceable to notebook observations or HAIDL components.
\subsection{Deriving Behavioral States}
\label{sec:behavioral_knowledge}
The classroom deployment provides an additional contextual observation:
whether \pegasusAI was expected to be available for a given
student--assignment pair, encoded as:
\begin{lstlisting}[style=bnf]
ExposureStatus ::=
      ExpectedUser
    | ExpectedNonUser
    | Control
    | CrossConditionUser
\end{lstlisting}
An \texttt{ExpectedUser} had assigned access and used \pegasusAI; an \texttt{ExpectedNonUser} had assigned access but did not use it; a \texttt{Control} had no assigned access and did not use the system; and a \texttt{CrossConditionUser} used it despite not having assigned access.
Thus, learning profile, exposure status, and interaction knowledge can now be combined.
\begin{lstlisting}[style=bnf]
LearningTrace ::=
      Learner
    | Assignment
    | ExposureStatus
    | LearningProfile
    | InteractionTrajectory
    | BehavioralState
BehavioralState ::=
      SelfSufficient
    | ExploratoryOverUser
    | DisengagedStruggler
    | ScaffoldedLearner
    | CrossConditionUser
\end{lstlisting}
For example, the analysis interprets an expected non-user with high performance as self-sufficient, whereas an expected non-user with low performance is interpreted as a disengaged struggler:
\[
{\scriptsize
\begin{gathered}
\texttt{ExpectedNonUser} + \texttt{HighPerformance} 
\Rightarrow_K \texttt{SelfSufficient}; \\ \\
\texttt{ExpectedNonUser} + \texttt{LowPerformance} 
\Rightarrow_K \texttt{DisengagedStruggler}
\end{gathered}
}
\]
For students expected to use \pegasusAI, interaction patterns further
refine interpretation. Low performance paired with productive
help-seeking---long studied in intelligent tutoring
systems~\cite{aleven2003help}---or sustained engagement indicates a
scaffolded learner, where \texttt{ProductiveHelp} aggregates error-recovery, next-step, and conceptual-help interactions:
\[
{\scriptsize
\begin{gathered}
\texttt{ExpectedUser} + \texttt{LowPerformance} + \\
(\texttt{ProductiveHelp} \lor \texttt{HighEngagement}) 
\Rightarrow_K \texttt{ScaffoldedLearner}
\end{gathered}
}
\]
High performance together with comparatively high AI use dominated by validation-oriented interactions evidences exploratory over-use, while cross-condition use is retained explicitly rather than interpreted as a normal help-seeking persona:
\[
{\scriptsize
\begin{gathered}
\texttt{ExpectedUser} + \texttt{HighPerformance} 
{}+ \texttt{HighBotUse} + \texttt{ValidationDominant} \\
\Rightarrow_K \texttt{ExploratoryOverUser}; \\ \\
\texttt{CrossConditionUse} \Rightarrow_K \texttt{CrossConditionUser}
\end{gathered}
}
\]
These behavioral states are not diagnoses or intrinsic student traits.
They are reproducible interpretations of observable behavior for a
particular student--assignment pair.
\subsection{What the Language Reveals}
The resulting hierarchy is: 
\[
{\scriptsize
\begin{gathered}
\text{Observations} \rightarrow \text{Interactions} \rightarrow \text{Events} 
\rightarrow \text{Trajectories} \rightarrow \text{Behavioral Knowledge}
\end{gathered}
}
\]
capturing the distinction between storing interaction data and reasoning
over it: a runtime error is an observation; an AI request made in that
context is error-recovery behavior; multiple such events form a
trajectory; and combined with execution-derived performance, engagement,
and experimental context, a trajectory supports a behavioral
interpretation. Conventional chat histories preserve text exchanges but
lack a common semantic representation connecting the request to
computational state, errors, task context, feedback, and subsequent
interactions; HAIDL makes these relationships explicit and derivable
through inspectable rules. The classroom deployment demonstrates this
capability, but the mechanism is not inherently educational---the same
domain-specific state and derivation rules could instead express
interaction knowledge about scientific workflows, software debugging, or
other forms of human--AI collaboration. HAIDL's contribution is thus a
compositional language, not a fixed taxonomy of learning behaviors.
\section{Case Study: Deriving Interaction Knowledge}
\label{sec:case_study}
We apply the Section~\ref{sec:knowledge} framework to the classroom deployment in Section~\ref{sec:deployment}, covering 140 student--assignment pairs from 35 students across four assignments and combining notebook execution data, \pegasusAI interaction logs, and access conditions. Our goal is not to estimate \pegasusAI's causal effect on performance, but to test whether first-class interaction data let us distinguish forms of human--AI behavior that are ambiguous from notebook traces or AI logs alone.
\subsection{Interaction Behavior and Learning Profiles}
\label{sec:case_interactions}
We first examine interaction events derived from \pegasusAI records, classifying each event's role (Section~\ref{sec:interaction_events}) from the source cell, execution state, errors, and request--response pair; we then examine notebook-derived learning profiles from performance and engagement proxies. Table~\ref{tab:events_profiles} reports both distributions by assignment.
\begin{table*}[t]
\centering
\caption{Derived interaction events and notebook-derived learning profiles by assignment.}
\label{tab:events_profiles}
\scriptsize
\begin{tabular}{lrrrrrrrr}
\toprule
& \multicolumn{4}{c}{\textbf{Interaction Events}} & \multicolumn{4}{c}{\textbf{Learning Profiles}} \\
\cmidrule(lr){2-5} \cmidrule(lr){6-9}
\textbf{Assign.} & \textbf{Error} & \textbf{Stuck/Next} & \textbf{Concept.} & \textbf{Valid.} & \textbf{HPHE} & \textbf{HPLE} & \textbf{LPHE} & \textbf{LPLE} \\
\midrule
03 & 44  & 67 & 2   & 21 & 19 & 0 & 0 & 16 \\
04 & 106 & 0  & 163 & 14 & 18 & 0 & 0 & 17 \\
07 & 31  & 37 & 5   & 0  & 22 & 0 & 8 & 5  \\
08 & 150 & 13 & 119 & 3  & 19 & 0 & 6 & 10 \\
\bottomrule
\end{tabular}
\end{table*}
Interaction events varied substantially across assignments. Assignment~03 (algorithmic problem solving) generated the most stuck/next-step, error-recovery, and validation requests. Assignment~04 shifted toward conceptual help and error recovery, while PySpark Assignments~07 ($k$-means) and 08 (a multi-stage MLlib pipeline) were dominated by error recovery, with Assignment~08 also generating substantial conceptual help. Thus, a simple ``number of prompts'' metric obscures important differences: the same \pegasusAI call can serve different roles depending on the task and execution state. HAIDL captures these differences by linking each request to its execution context.

The learning profiles add important nuance. The first two assignments are dominated by High-Performance/High-Engagement (HPHE) and Low-Performance/Low-Engagement (LPLE), while the Spark assignments introduce Low-Performance/High-Engagement (LPHE) students (8 pairs in Assignment~07 and 6 in Assignment~08), with no HPLE cases. LPHE captures \emph{engaged struggle}: students who touched many cells but completed few. In contrast, LPLE combines low performance with low activity. Interaction counts alone reveal little about outcomes: across all 140 pairs, interaction count has only weak correlations with successfully executed cells ($r=0.12$) and touched cells ($r=0.12$). Thus, frequency alone does not indicate whether \pegasusAI helped or hindered performance; events and learning profiles instead provide complementary views of \emph{how} AI was used.
\subsection{Exposure and Derived Behavioral States}
\label{sec:case_exposure}
The crossover design separates availability from actual use; combining it with learning profile and interaction behavior (Section~\ref{sec:behavioral_knowledge}) yields the behavioral states below. Table~\ref{tab:exposure_persona} reports both distributions by assignment; Cross-Condition is listed once, since a cross-condition user's exposure and behavioral-state counts coincide by definition.
\begin{table*}[t]
\centering
\caption{\pegasusAI exposure states and derived behavioral states by assignment.}
\label{tab:exposure_persona}
\scriptsize
\begin{tabular}{lrrrrrrrr}
\toprule
& \multicolumn{3}{c}{\textbf{Exposure States}} & \multicolumn{5}{c}{\textbf{Behavioral States}} \\
\cmidrule(lr){2-4} \cmidrule(lr){5-9}
\textbf{Assignment} & \textbf{Exp. User} & \textbf{Exp. Non-User} & \textbf{Control} & \textbf{Self-suff.} & \textbf{Diseng.} & \textbf{Scaffolded} & \textbf{Expl. Over.} & \textbf{Cross-Cond.} \\
\midrule
03 & 9 & 8  & 12 & 4 & 4 & 8 & 1 & 6  \\
04 & 9 & 9  & 7  & 7 & 3 & 8 & 0 & 10 \\
07 & 7 & 10 & 14 & 5 & 5 & 7 & 0 & 4  \\
08 & 8 & 10 & 11 & 7 & 5 & 6 & 0 & 6  \\
\bottomrule
\end{tabular}
\end{table*}
Two exposure results stand out. Assigned access did not guarantee use: 37 student--assignment pairs had access but no interactions, so non-use cannot be equated with unavailability. Conversely, 26 pairs used \pegasusAI outside their assigned-access condition---occurring in every assignment, largest in Assignment~04 (10 cases)---undermining a simple treatment-versus-control reading; we keep these separate rather than folding them into expected-user or control. An interaction log alone cannot show whether use was expected under the design, nor distinguish a control student from one who simply did not seek help; only combining HAIDL records with experimental context resolves these cases.
The behavioral states follow the same logic. \texttt{ScaffoldedLearner} appears in all four assignments (29 pairs total)---students with need or difficulty who used \pegasusAI for debugging, next-step, or conceptual support, evidence many used the assistant as intended. \texttt{DisengagedStruggler} also appears in every assignment (three to five cases each): students with access and weak notebook outcomes but little or no \pegasusAI use, indistinguishable from self-sufficient students in the AI log alone but separated by notebook evidence. Self-sufficient students combine strong notebook performance with non-use, indicating independent progress rather than hidden struggle. Exploratory over-use appears only once, with no consistent \texttt{AnswerExtractor} cases, so this deployment mixes productive scaffolding, independent work, and unmet need rather than indiscriminate AI dependence. The 26 cross-condition pairs are analyzed separately, since the design does not support treating them as typical assigned-access help seeking.
\subsection{What the Derived Knowledge Reveals}
\label{sec:case_findings}
Analyzing interaction data and notebook execution evidence separately, then combining them, is more informative than either alone, in four ways.
\textbf{Interaction counts conceal how AI is used}: \pegasusAI's role shifts with the task---next-step help and validation for Assignment~03, error recovery for Assignment~07, conceptual and error recovery for Assignments~04 and~08---a distinction simple request counts cannot reveal.
\textbf{Notebook traces distinguish forms of struggle}: the Spark assignments surface LPHE students who keep attempting cells despite limited success, an engaged-struggle state distinct from LPLE, with HAIDL records showing whether they seek debugging, conceptual, or next-step help.
\textbf{Zero AI interactions are ambiguous}: a non-user with strong performance suggests self-sufficiency, a non-user with weak performance a disengaged struggler---identical states in the AI log alone.
\textbf{Observed use and experimental exposure are not equivalent}: the 26 cross-condition cases show a simple used/did-not-use variable cannot capture real deployment, since HAIDL records observed behavior while the access schedule supplies context to interpret it.
These results do not show that \pegasusAI improves learning, nor is that their purpose; they support this paper's central claim that structured interaction data, combined with computational and experimental evidence, yields knowledge none of these sources provides alone---every behavioral state traces back to specific HAIDL fields, notebook observations, and derivation rules, the practical value of treating human–AI interactions as first-class data.
\section{Cost of First-Class Interaction Capture}
\label{sec:capture_cost}
Treating human–AI interactions as first-class data requires \pegasusAI to capture computational context, build a HAIDL object, store it, and retrieve its components for analysis---steps adding runtime and storage overhead beyond the model invocation. This section's goal is narrow: measuring that overhead, not benchmarking providers or model quality.
\subsection{Capture Overhead}
\label{sec:capture_overhead}
\begin{figure*}[tp]
    \centering
    \subfloat[Azure: o4-mini]{\includegraphics[width=0.30\textwidth]{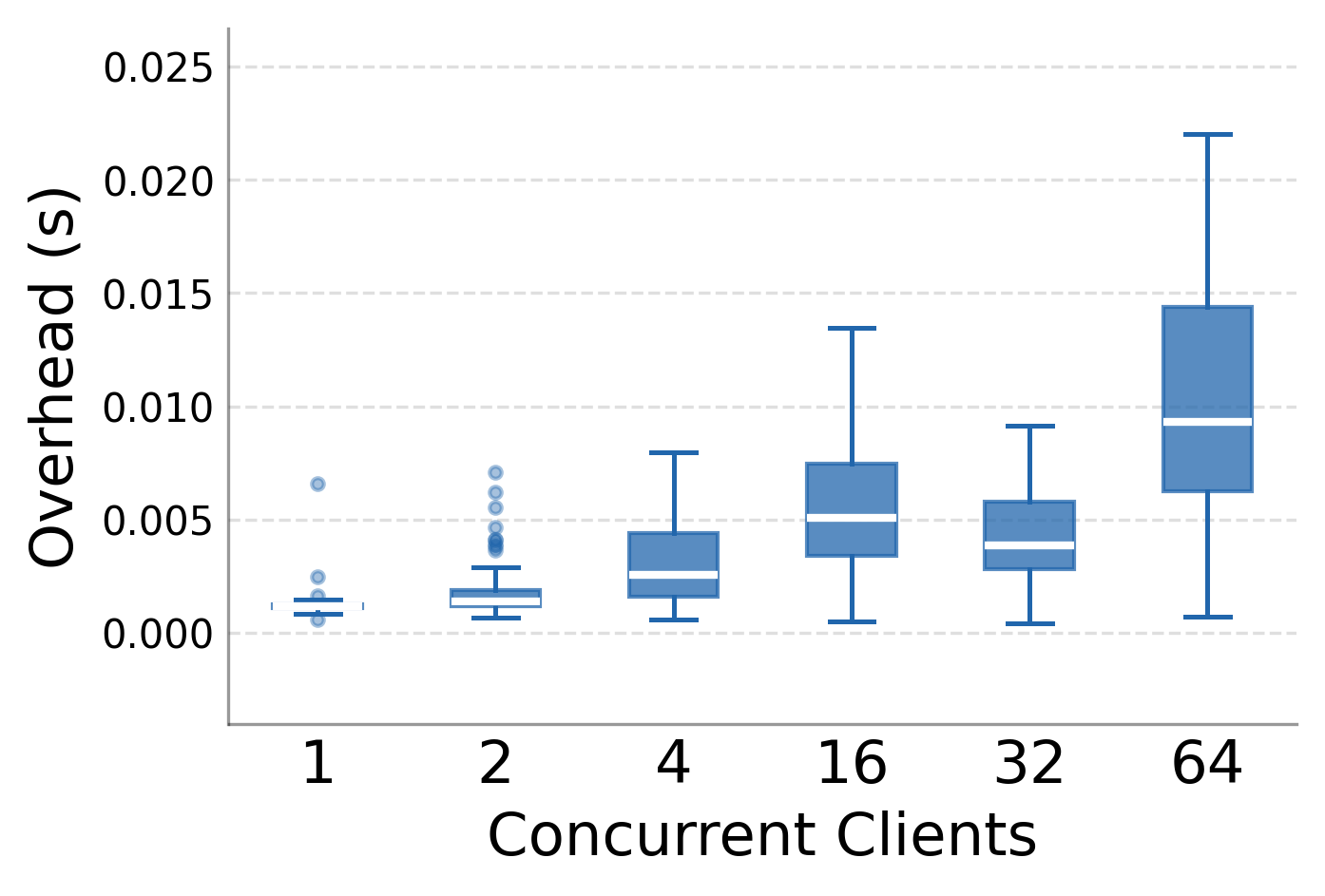}\label{fig:azure_o4-mini}}
    \hfill
    \subfloat[vLLM: ministral-3-14b]{\includegraphics[width=0.30\textwidth]{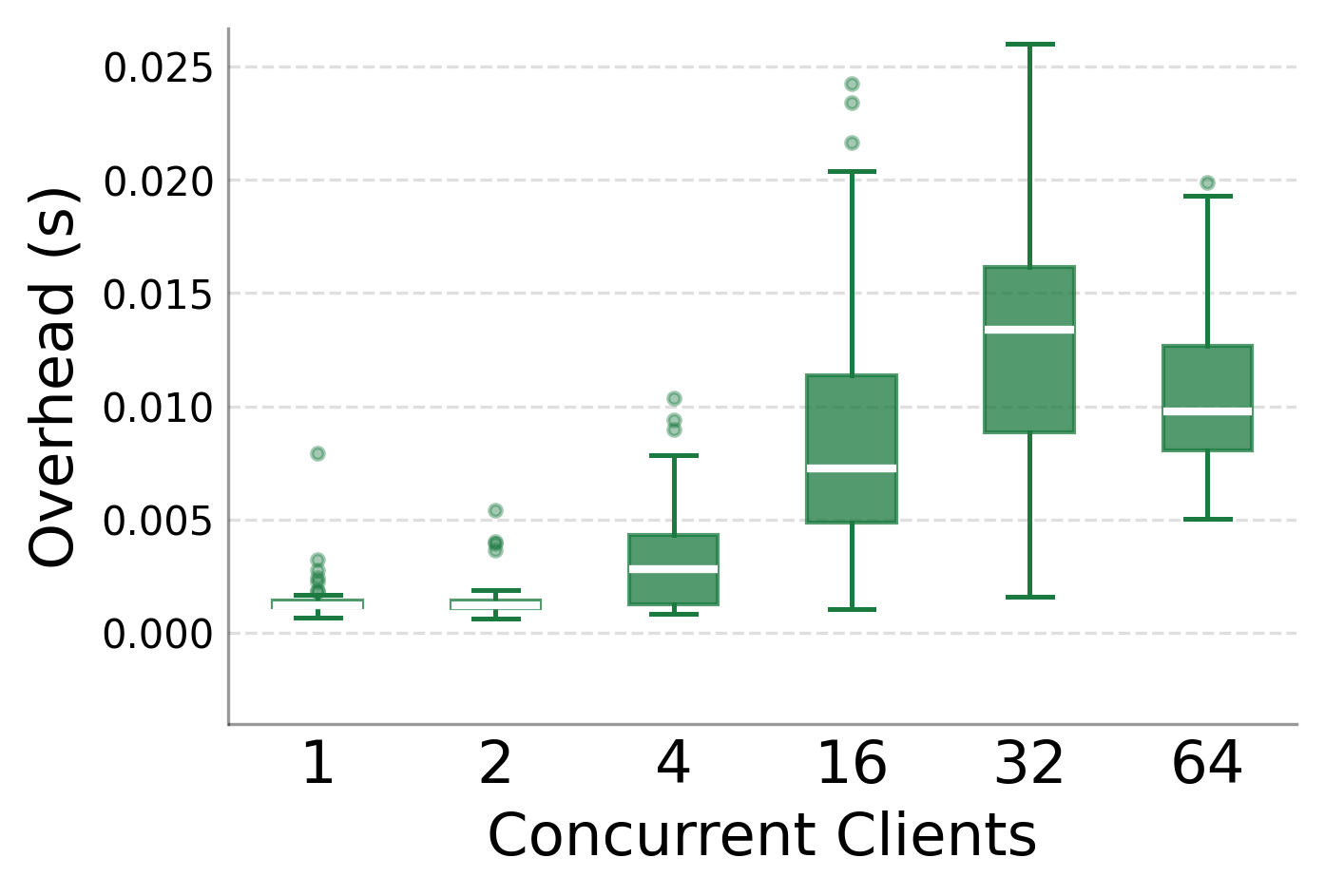}\label{fig:vllm-ministral}}
    \hfill
    \subfloat[Sambanova: gemma-4-31B-it]{\includegraphics[width=0.30\textwidth]{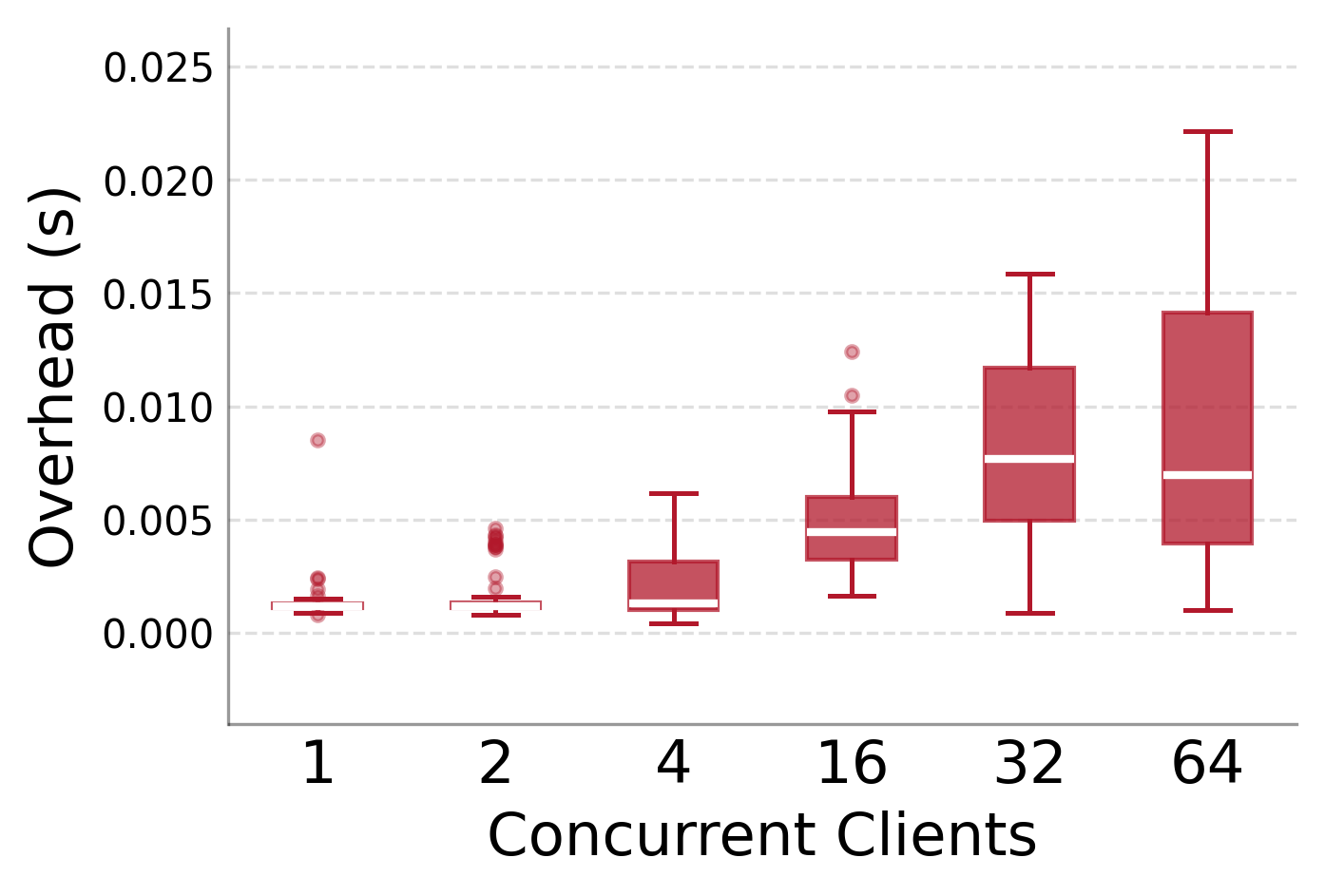}\label{fig:sambanova-gemma}}
    \caption{Per-model overhead distributions across concurrent clients and three providers (Azure, vLLM, Sambanova), showing consistently low, provider-agnostic overhead.}
    \label{fig:results_overhead}
\end{figure*}

For each \pegasusAI interaction, we measure two latencies: the server-side request latency $T_{\mathrm{server}}$, spanning the complete request callback---collecting notebook state, building the request, invoking the LLM, assembling the HAIDL object, and persisting it---and the LLM invocation latency $T_{\mathrm{LLM}}$ alone. The additional cost attributable to first-class capture is therefore
\begin{equation}
Overhead =
T_{\mathrm{server}} - T_{\mathrm{LLM}}
\end{equation}
i.e., the processing \pegasusAI adds beyond the LLM call itself. Varying the AI provider, LLM model, and number of concurrent client requests, Figure~\ref{fig:results_overhead} shows overhead growing steadily with concurrency yet staying within 25ms across all configurations---confirming our runtime introduces negligible, provider-agnostic overhead and remains practical for interactive notebook use.
\subsection{Persistence Cost}
\label{sec:persistence_cost}
Persisting an interaction as a HAIDL object incurs storage and database-write costs, with object size depending on the amount of source code, execution output, traceback information, prompt context, AI response, and metadata. Table~\ref{tab:persistence_characteristics} reports these sizes and persistence times: object sizes vary substantially (median 4.53~KB, P95 61.79~KB), yet persistence latency remains low (median 1.40~ms, P95 1.97~ms) even for the largest objects.
\subsection{Retrieval Cost}
\label{sec:retrieval_cost}
First-class interaction data are useful only if stored representations can be efficiently retrieved. We evaluate three queries corresponding to the derivations in Sections~\ref{sec:knowledge} and~\ref{sec:case_study}: (1)~all interactions for a student and assignment, (2)~interactions containing an \texttt{ErrorOutput} (evidence for \texttt{ErrorRecovery}), and (3)~temporally ordered interactions for an \texttt{InteractionTrajectory}. As Table~\ref{tab:persistence_characteristics} shows, all three have similar, low retrieval cost (median $\approx$0.4~ms, P95 below 0.8~ms).
\begin{table}[t]
\centering
\caption{Persistence and retrieval characteristics of HAIDL objects and queries.}
\label{tab:persistence_characteristics}
\scriptsize
\begin{tabular}{lrr}
\toprule
\textbf{Measure} & \textbf{Median} & \textbf{P95} \\
\midrule
HAIDL object size (KB)              & 4.53  & 61.79 \\
Persistence latency (ms)            & 1.40  & 1.97  \\
Retrieval: student--assignment (ms) & 0.399 & 0.581 \\
Retrieval: error-containing (ms)    & 0.399 & 0.711 \\
Retrieval: ordered trajectory (ms)  & 0.401 & 0.652 \\
\bottomrule
\end{tabular}
\end{table}
\subsection{Discussion}
\label{sec:capture_discussion}
The key systems question is not whether \pegasusAI outperforms an AI provider or database, but whether preserving full interaction context is practical for interactive use. These measurements quantify that cost against the benefit shown in Section~\ref{sec:case_study}: preserving computational state, context, provenance, and temporal structure enables knowledge otherwise hard to reconstruct. First-class capture thus trades measurable systems cost for a persistent data product supporting querying, reconstruction, and knowledge derivation.
\section{Related Work}
\label{sec:related_work}
Table~\ref{tab:related_comparison} compares HAIDL against representative
systems from three related families; none couples both sides of a
human--AI exchange or defines derivation semantics for the coupled
record, as detailed below.
\begin{table}[t]
\centering
\caption{What existing representations capture vs.\ HAIDL. Columns:
AI observability (OTel GenAI~\cite{opentelemetrygenai}), agent-centric
provenance (PROV-AGENT~\cite{prov-agent}), and programming-process data
(ProgSnap2~\cite{progsnap2}).}
\label{tab:related_comparison}
\scriptsize
\setlength{\tabcolsep}{3.5pt}
\begin{tabular}{p{3.35cm}cccc}
\toprule
\textbf{Captured information} &
\textbf{OTel} &
\textbf{PROV-} &
\textbf{Prog-} &
\textbf{HAIDL} \\
&
\textbf{GenAI} &
\textbf{AGENT} &
\textbf{Snap2} &
\\
\midrule
Prompt, response, model metadata            & \checkmark & \checkmark & --- & \checkmark \\
Computational state motivating a request    & --- & \checkmark$^{\ddagger}$ & \checkmark$^{\dagger}$ & \checkmark \\
User assessment of the response             & --- & --- & --- & \checkmark \\
User actions after the response             & --- & --- & \checkmark$^{\dagger}$ & \checkmark \\
Interaction as an explicit state transition & --- & --- & --- & \checkmark \\
Formal grammar with derivation semantics    & --- & --- & --- & \checkmark \\
\bottomrule
\end{tabular}
\vspace{2pt}
\begin{minipage}{\linewidth}
\scriptsize
$^{\dagger}$\,Captures code states and executions, but has no notion
of AI assistance.\\
$^{\ddagger}$\,Captures workflow-side context for agent decisions,
not the user state that motivates a request.
\end{minipage}
\end{table}
\subsection{AI Assistants and Programming Support}
AI assistants increasingly support programming through code generation, debugging, and explanation~\cite{ziegler2022productivity,barke2023grounded}, with effective use depending on context, response interpretation, and validation~\cite{vaithilingam2022expectation,prather2023s,nguyen2024beginning,scholl2024novice}, including in classroom and notebook deployments~\cite{kazemitabaar2023novices,kazemitabaar2024codeaid,mcnutt2023notebooks}. Related work infers usage patterns from conversation logs~\cite{llm-intents}, IDE telemetry~\cite{mozannar2024cups}, and large-scale corpora~\cite{zhao2024wildchat}, but these are not linked to the computational state motivating each request---reproducing the ambiguities of Section~\ref{sec:introduction}. HAIDL instead couples request, response, computational context, and assessment in one durable object, turning per-study analysis into queries over a standard representation.
\subsection{Provenance and AI Observability}
Provenance systems record processes, dependencies, and execution states for reproducibility~\cite{davidson_provenance,oliveira_provenance,pegasus-wms}, built on the W3C PROV model~\cite{w3c-prov} and extended by ProvONE~\cite{cuevas_provenance}, notebook-provenance systems~\cite{pimentel2015provenance}, and Workflow Run RO-Crate~\cite{leo2024workflowrunrocrate}. Closest to our setting, PROV-AGENT~\cite{prov-agent} represents AI-agent prompts and decisions as provenance nodes, LLMography~\cite{llmography} derives auditability indicators from LLM conversations, and OpenTelemetry GenAI and OpenInference standardize model-invocation telemetry~\cite{opentelemetrygenai,openinference}. This family keys to the model invocation or agent actions, not the user's computational state or assessment (Table~\ref{tab:related_comparison}); HAIDL contributes the human-side coupling and derivation semantics ($\Rightarrow_K$) these models omit.
\subsection{Interaction Data Standards and Formal Models}
Interoperable activity standards---ProgSnap2 for programming-process data~\cite{progsnap2}, the Experience API~\cite{kevan2016experience}, and profile-based event models~\cite{scala2014}---predate AI assistance and provide no rules for deriving higher-level constructs from recorded events; such analyses are re-implemented per study. Grammar-based formalisms have separately been applied to constrained generation~\cite{wang2023grammar_prompting} and workflow resource-allocation~\cite{estrada2015genetic}. HAIDL applies the same principle, pairing a BNF-inspired grammar with an explicit derivation relation these standards lack (Section~\ref{sec:knowledge}).
\section{Conclusion}
\label{sec:conclusion}
Human–AI interactions are central to computational and scientific workflows, yet they are still logged only as transient prompts and responses, discarding the computational state, task context, feedback, provenance, and timing needed to understand the collaboration. This paper introduces the Human–AI Interaction Description Language (HAIDL), a formal language that treats each interaction as first-class data: a structured object with composition and derivation semantics relating these objects across states and over time. \pegasusAI instantiates HAIDL in JupyterLab, capturing interaction context, AI responses, feedback, and provenance during live computational work.
A classroom deployment demonstrates the value of this representation. HAIDL objects compose into semantic interaction events---error recovery, next-step assistance, conceptual help, validation---and, combined with notebook execution and experimental context, into higher-level behaviors distinguishing independent progress, scaffolded AI use, and disengaged struggle: patterns invisible from interaction counts or notebook traces alone. These results are not a causal claim about AI's effect on learning; they show structured interaction data supports reproducible, evidence-traceable interpretation.
The broader contribution is not a classroom taxonomy but a data model and language for preserving human–AI collaboration in queryable, composable form. The same primitives extend to scientific workflows, software development, and AI-assisted experimentation wherever understanding an AI collaborator's role requires more than logging what was said---it requires preserving the state, provenance, and consequences of each exchange. Treating interactions as first-class data is what makes human–AI collaboration inspectable, reproducible, and ultimately explainable.

\section*{Acknowledgments}
This work was supported by National Science Foundation Awards \#2513101, \#2530461, \#2449103, \#2331152, and \#2334945. This work used resources from NAIRR \#260183. 

{\let\footnotesize\scriptsize
\renewcommand{\baselinestretch}{0.95}
\bibliographystyle{IEEEtran}
\bibliography{references}
}
\end{document}